\RequirePackage{letltxmacro}
\LetLtxMacro{\LaTeXtextbf}{\textbf}
 \documentclass{vgtc}
 \LetLtxMacro{\textbf}{\LaTeXtextbf}
\usepackage{cite}
\usepackage{amsmath,amssymb,amsfonts}
\usepackage{graphicx}
\usepackage{textcomp}

\usepackage{booktabs}
\usepackage{hyphenat}
\usepackage{algorithm2e}
\usepackage{hyperref}
\usepackage{dblfloatfix}

\usepackage{color,soul}

\SetCommentSty{}

\def\BibTeX{{\rm B\kern-.05em{\sc i\kern-.025em b}\kern-.08em
    T\kern-.1667em\lower.7ex\hbox{E}\kern-.125emX}}

\title{Lifelong Continual Learning for Anomaly Detection: New Challenges, Perspectives, and Insights\thanks{The paper was published in IEEE Access journal. DOI: 10.1109/ACCESS.2024.3377690}}
\author{Kamil Faber \\ \scriptsize{AGH University of Krakow}  \\ \scriptsize{Poland} \\ \scriptsize{kfaber@agh.edu.pl} 
\and Roberto Corizzo \\ \scriptsize{AGH University of Krakow} \\ \scriptsize{American University} \\ 
\scriptsize{Washington, DC} \\ \scriptsize{rcorizzo@american.edu} \and
Bartlomiej Sniezynski \\ \scriptsize{AGH University of Krakow} \\  \scriptsize{Poland} \\ \scriptsize{bartlomiej.sniezynski@agh.edu.pl} \and 
Nathalie Japkowicz \\ \scriptsize{American University} \\ 
\scriptsize{Washington, DC} \\ \scriptsize{japkowic@american.edu}
}

\abstract{
Anomaly detection is of paramount importance in many real-world domains characterized by evolving behavior, {such as monitoring cyber-physical systems, human conditions and network traffic.} 
{Current research in anomaly detection leverages offline learning working with static data or online learning focusing on constant adaptation to evolving data.}
{At the same time,} lifelong learning represents an emerging trend, answering the need for machine learning models that continuously adapt to new challenges in dynamic environments while retaining past knowledge. 
{Although this aspect could be beneficial to build effective and robust anomaly detection models, lifelong learning research is mainly dedicated to proposing new model update strategies
in image classification and reinforcement learning domains.}
{The limited scope addressed by lifelong learning works thus far creates a gap in understanding whether such techniques and capabilities can be fruitfully exploited in  anomaly detection contexts, which represents the main motivation of this paper.}
{More specifically, anomaly detection provides unique challenges, such as an evolving normal class and limited availability of anomalies, which significantly differs from the landscape and scenarios of lifelong image classification and reinforcement learning.} 
In this paper, we face this issue by exploring, motivating, and discussing lifelong anomaly detection, as well as providing foundations with regard to scenarios, strategies, and metrics.
First, we explain why lifelong anomaly detection is relevant, defining challenges and opportunities to design anomaly detection methods that deal with lifelong learning complexities.
Second, we  {formulate and characterize lifelong learning settings tailored for anomaly detection problems, and design} a scenario generation procedure that enables researchers to experiment with lifelong anomaly detection using existing datasets.
Third, we perform experiments with popular anomaly detection methods on proposed lifelong scenarios, emphasizing the gap in performance that could be filled with the adoption of lifelong learning.
{In summary, our efforts are directed at assessing the performance of non-lifelong anomaly detection models in lifelong scenarios and how the adoption of lifelong learning impacts their learning capabilities.}
Overall, we conclude that the adoption of lifelong anomaly detection is important to design more robust models that provide a comprehensive view of the environment, as well as simultaneous adaptation and knowledge retention.}

\begin{document}
\maketitle


\section{Introduction}
\label{sec:introduction}
\setlength{\textfloatsep}{6pt}


Anomaly detection is the task of finding {anomalous data instances that represent a} deviation from the normal conditions of a process \cite{aggarwal2017introduction,schmidl2022anomaly}. The capability to detect anomalous behavior is of paramount importance in many disciplines and real-world applications, such as intrusions in network traffic \cite{faber2021autoencoder}, irregular behavior in cyber-physical systems such as smart grids \cite{corizzo2021spatially}, {as well as IoT environments {\cite{fahim2019anomaly}},} or defects in manufacturing processes \cite{ALFEO2020272}. 
%
The most widespread approach in machine learning is to model the normal behavior of the system and identify anomalies as data instances that significantly differ from the modeled behavior \cite{pang2021deep}. This choice reflects the limited availability of anomalies compared to the large availability of normal data, which results in the inability to model the anomaly class accurately. Moreover, it responds to the necessity of detecting anomalies with varying morphology unknown at training time \cite{chalapathy2019deep,10.1145/3439950}.  

Most 
works in anomaly detection deal with the problem in an offline (batch) or online (stream) manner \cite{aggarwal2017introduction}. 
In evolving environments, models will become outdated and require updates, either as a full retraining stage (batch models), or as an online adaptation stage following concept drift detection (stream models) \cite{fenza2019}. 

{

Both types of approaches are equally valid depending on the domain characteristics. For instance, offline approaches are commonly used for tasks such as lesion detection in medical images \cite{9618843} and gravitational waves detection \cite{corizzo2020scalable}.
On the other hand, online approaches are common in domains characterized by a temporal dimension, such as real-time fatigue detection \cite{laxhammar2013online} and crowd anomaly detection \cite{FENG2017548}.  
Updating models allows them to 
adapt to the changing conditions of the normal class. However, it is noteworthy that updating the model has the side effect of gradually leading to forgetting past knowledge \cite{PARISI201954,cossu2021continual}.
}
Forgetting is a widely known phenomenon in data streams and online learning, and it is considered to be a positive feature in some scenarios as it allows models to focus on the most recent data characteristics \cite{gama2010knowledge}. 
{
For instance, in crowd anomaly detection \cite{FENG2017548}, forgetting is suitable since it is assumed that only the people present in the current monitored environment (i.e., the most recent data) are the relevant ones to predict anomalies within that particular environment and at that particular time.
}

On the other hand, lifelong continual machine learning \footnote{Terms ''lifelong'' and ''continual'' are used interchangeably in existing literature. From now on, we will use the term lifelong to refer to this learning setting.} research shows that forgetting is a problem that negatively affects models' performance when previously experienced conditions reoccur in the future \cite{parisi2018role, kirkpatrick2017overcoming,10341211,liu2021overcoming}.  
For this reason, lifelong learning seeks to find a balance between adapting to new knowledge while retaining past knowledge, inspired by 
biology, neuroscience, and computer science \cite{PARISI201954,chaudhry2021using,masana2022class}.
{
For instance, a popular example in lifelong reinforcement learning is having an agent able to learn how to play new games while not losing the ability to play previously known ones. 
}

Despite the emerging interest in lifelong learning, current research on this topic is mainly focused on computer vision and reinforcement learning \cite{PARISI201954, abel2018policy,parmar2023open,baker2023domain}{, with new trends including bridging active learning with open world learning \mbox{\cite{mundt2023wholistic}} and applications to robotics \mbox{\cite{10251045}}, as well as online image classification \mbox{\cite{mai2022online}}. In contrast,} anomaly detection problems are still poorly explored. 
In this paper, we focus on lifelong learning from an anomaly detection perspective, showing that lifelong learning capabilities can bring several advantages in many real-world settings. We argue that {considering them} will yield more sophisticated models that can detect anomalies while adapting to changing environments and avoiding forgetting knowledge acquired in the past.
One of the domains where this capability is crucial is cybersecurity {\mbox{\cite{faber2023collaborative}}}. For example, monitoring network traffic has to deal with dynamic conditions such as changes in the infrastructure, user behaviors, as well as new types of traffic and protocols {\mbox{\cite{javaheri2023fuzzy}}}.
Other examples of domains that we further describe in the paper include human condition monitoring and fault detection in industrial settings, although many more can be drawn.  

%
Moreover, we leverage lessons learned in recent lifelong machine learning studies to showcase the current limitations of anomaly detection methods when exposed to lifelong learning scenarios. To this end, we formalize lifelong anomaly detection and devise the desiderata of models and scenarios, building the foundations for future work. 

\begin{figure}[h!]
    \centering
    \includegraphics[width=0.47\textwidth]{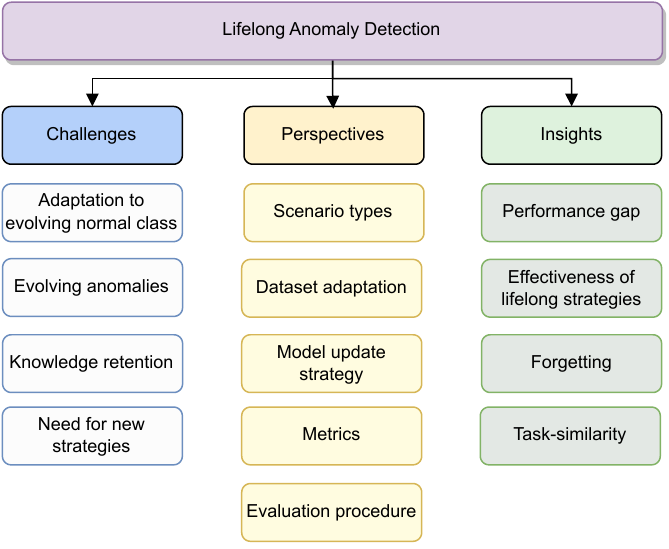}
    \caption{{General view of lifelong learning in terms of challenges (see Section {\ref{sec:lifelong_anomaly_detection}}), perspectives (see Section {\ref{sec:scenarios}}), and insights (see Section {\ref{sec:discussion}}) examined in our paper. 
    }}
    \label{fig:general}
\end{figure}

{Our analytical scope focuses on two main research questions: }
\begin{itemize}
    \item {\textbf{RQ1:} Do lifelong scenarios impact the performance of non-lifelong anomaly detection models?}
    \item {\textbf{RQ2:} Does the adoption of knowledge retention capabilities of lifelong learning provide a valuable improvement in the learning capabilities of existing anomaly detection models in complex lifelong scenarios?}
\end{itemize}
{With the first research question, we aim to assess whether current anomaly detection models are effective in lifelong learning scenarios and if there is a gap that needs to be addressed. With the second question, we aim to verify whether adopting lifelong learning strategies can be beneficial in anomaly detection contexts. 
}

The contributions of this study can be summarized as follows:
\begin{itemize}
    \item{Bridging the gap between anomaly detection and lifelong learning, presenting the benefits of lifelong anomaly detection over conventional anomaly detection, which paves the way for new methods that are more robust in real-world domains;
    } 

    \item{Devising a characterization of anomaly detection scenarios from a lifelong learning perspective, which sheds light on how to design and build more challenging benchmarks for anomaly detection;}

    \item{An open-source\footnote{\url{https://github.com/lifelonglab/lifelong-anomaly-detection-scenarios}} implementation for scenario generation method, which can be applied to any anomaly detection dataset, facilitating wider adoption of lifelong learning in anomaly detection settings.
    }
    \item{Evaluating popular anomaly detection methods on our proposed lifelong scenarios, emphasizing challenges and limitations that occur in this setting.}

\end{itemize}

{
We summarize challenges, perspectives, and insights in Figure {\ref{fig:general}}.
Our contributions allow us to highlight the potential of lifelong anomaly detection as a new, promising research direction. The scenarios we designed and the experimental results we obtained in our study help us showcase new challenges, perspectives, and insights brought by lifelong learning research in the context of anomaly detection. 
By doing so, we aim to increase awareness of the potential of lifelong anomaly detection while rationalizing and providing foundations with regard to scenarios, strategies, and metrics. Our study aims to streamline the adoption of lifelong anomaly detection for researchers and practitioners.
}

The paper is structured as follows. 
Section \ref{sec:background} summarizes related works in lifelong learning and anomaly detection. Section \ref{sec:lifelong_anomaly_detection} explores the transition from conventional to lifelong anomaly detection. Section \ref{sec:scenarios} describes the proposed lifelong scenario design procedure. Section \ref{sec:experiments} discusses the experimental results obtained in our study. Finally, Section \ref{sec:conclusions} concludes the paper outlining directions of interest for future work.

\section{Background}
\label{sec:background}
In this section, we first briefly introduce lifelong learning along with the most popular types of methods. Second, we provide an overview of the current landscape of anomaly detection works. 

\subsection{Lifelong learning}
\label{sec:introduction_to_lifelong_learning}

Lifelong learning is a continuous process in which a series of different problems, defined as tasks, are presented to a learning method over time \cite{PARISI201954}.
{
In the most common lifelong learning settings, image classification, a few scenario types built upon task characterization have been proposed. The most popular ones are task-incremental, class-incremental, and domain-incremental {\cite{van2019three}}. 
Both class-incremental and task-incremental scenarios provide the model with new, previously unseen classes that need to be incorporated by the model {\cite{belouadah2021comprehensive,de2021continual}}. On the other hand, domain-incremental scenarios provide new distributions of already known classes {\cite{10.1007/978-3-031-18840-4_7}}.
Emerging trends involve online lifelong learning {\cite{De_Lange_2021_ICCV,carta2023comprehensive}}, simultaneous adaptation in terms of new instances and new classes {\cite{lomonaco2019nicv2}}, and repetition of observed tasks {\cite{cossu2022class}}.
}

In the lifelong setting, the general goal for the learner is to be able to pick up new skills, adjust to newly presented tasks, and draw on previously learned information to tackle both new obstacles and the recurrence of previously seen tasks \cite{9722932}.
The key difference between lifelong learning and incremental/online learning is that, in lifelong learning, the attention is not solely focused on adaptation but also on knowledge retention and a model's ability to simultaneously handle all tasks, avoiding forgetting \cite{kirkpatrick2017overcoming}.  
{
To this end, lifelong learning strategies are inspired by diverse disciplines, including neuroscience and biology \cite{PARISI201954}.

Lifelong learning approaches proposed thus far fall into three main categories.

}

\textbf{Regularization-based strategies} work by introducing constraints on weight updates during the incremental training of neural networks. 
One of the initial ideas is to prevent updates for weights learned on previously trained tasks \cite{razavian6382cnn}.
{

Another approach is to freeze the first layers in the model architecture to mitigate forgetting previous tasks while leaving the last fully connected layer unfrozen to be updated with new tasks \cite{lomonaco2017core50}.
On the other hand, methods such as EWC  \cite{kirkpatrick2017overcoming} and LWF \cite{zhizhong2018}
modify the regularization loss using a knowledge distillation to prevent drastic changes in already learned weights, which are important to solve previous tasks. 
}

\textbf{Dynamic architectures strategies} adaptively manipulate the model architecture during the learning process. 
Methods usually expand the network by adding new neurons or layers as they encounter new tasks \cite{diethe2019continual}. 
{

More efficient methods augment dynamic adaptation with pruning capabilities, which keep model capacity under control by removing insignificant weights. 
Popular examples are PackNet \cite{mallya2017} and Winning Subnetworks \cite{kang2022}, which splits the model into independent sub-networks, each specialized in addressing a different task. This type of approach is also referred to as forget-free, which holds only under assumptions such as the availability of task labels and unlimited capacity. 
}

\textbf{Replay-based strategies} ensure that the knowledge from previously seen tasks is taken into consideration by the updated model by recurringly incorporating a summarized version of data from previous tasks in model updates. The most standard replay-based techniques focus on preserving knowledge by storing data samples from previously learned tasks in a memory buffer and replaying them during model update \cite{PARISI201954}. 
{

There are a few strategies devised to select the most relevant samples for every task, keep the replay buffer size compact, and, in turn, limit resource usage \cite{buzzega2021rethinking}.
The second category leverages generative models to generate artificial data samples from previous tasks each time the model is updated \cite{NIPS2017_0efbe980}. 
By doing so, generative replay eases the burden of storing data samples, reducing the impact of memory occupation that affects conventional replay strategies.
}

\subsection{Anomaly detection}

We now turn our attention to anomaly detection.
{As we noted in Section {\ref{sec:introduction}}, anomaly detection has become crucial for decision support in many domains. For instance, the work in {\cite{fahim2019anomaly}} emphasizes the importance of anomaly detection in IoT environments, such as transportation systems, health care systems, smart objects, and industrial systems.
Similarly, anomaly detection in log sequences is critical for ensuring operational and security integrity in heterogeneous systems {\cite{mvula2023heart}}.
Another interesting example is Industry 4.0, with an example of 3D printing {\cite{szydlo2021dataset}}, where early identification of malfunctions is fundamental to limit economic losses.
}

In addition to the online \textit{vs.} offline distinction mentioned in Section \ref{sec:introduction}, anomaly detection methods can also be characterized as supervised, semi\hyp supervised, and unsupervised \cite{pang2021deep,goldstein2016comparative}, based on data and labels availability.  
{

Supervised methods require labels for both normal and anomaly classes. They also need to be concerned about the class imbalance problem, and they are generally limited by the fact that they can only identify known anomalies rather than discover new ones.
Semi\hyp supervised methods are trained using exclusively normal data, and try to identify anomalies in unseen data based on their difference with respect to the learned data distribution of the normal class. 
%
Unsupervised methods are different in that they make no assumptions about labels in training data and are simply data-driven, i.e.{,} they fit the model based on all the available unlabelled data. 
Works in the literature \cite{goldstein2016comparative} suggest that semi\hyp supervised methods should be preferred if enough labeled normal data is available in order to achieve more robust models.

Due to the peculiarities of the learning settings, semi-supervised and unsupervised methods typically entail one-class learning models. 
Relevant examples of one-class learning anomaly detection methods are:
\textit{i)} Variational Autoencoder (VAE) \cite{kingma2013auto}, a neural-network reconstruction-based model with generative capabilities. The model is trained in a one-class manner by minimizing reconstruction error on training data, and the reconstruction error is used as an anomaly score; 
\textit{ii)} One-Class Support Vector Machine (OCSVM) \cite{williamson2000support}, which provides anomaly scores comparing new data with a hyperplane-based decision boundary learned during the training stage;
\textit{iii)} Local Outlier Factor (LOF) \cite{breunig2000kriegel}, which yields an anomaly score based on the ratio between the local density of new data samples with respect to the average local density of its nearest neighbors; \textit{iv)} Isolation Forest (IF) \cite{liu2008zhou}, which provides ensembles of trees and considers the length of the path from root to leaf to determine the anomaly score of new samples: a shorter (or longer) path means that a data point is more (or less) likely to be an anomaly; \textit{v)} Copula-based anomaly detection (COPOD) \cite{li2020copod}, which predicts the degree of ``extremeness" of data samples based on tail probabilities of an empirical copula, a multivariate cumulative distribution function. 
}

However, while these methods are well-established and perform well in a wide number of scenarios, they do not provide simultaneous knowledge retention and model adaptation, lacking lifelong capabilities.
We can observe that recent research works started addressing lifelong anomaly detection. Examples include the adoption of meta-learning to estimate parameters for multiple tasks in one-class image classification \cite{frikha2021arcade}, transfer learning in video anomaly detection \cite{doshi2020continual}, change-point detection coupled with memory organization \cite{corizzo2022cpdga,FABER2023248}{, and leveraging user feedback to improve model performance {\cite{faber2022active,du2019lifelong}}.}

{
Despite the clear advantages that lifelong learning could provide in anomaly detection methods, the number of published works is still rather limited.
We attribute this scarcity to the novel and emerging nature of the subject and to the lack of established practices, protocols, and guidelines. 
Our study attempts to fill this gap by increasing the awareness of the potential of lifelong anomaly detection, while rationalizing and providing foundations with {regard} to scenarios, strategies, and metrics, which foster a simplified adoption of the lifelong learning framework for researchers and practitioners. 
}

\section{From Anomaly Detection to Lifelong Anomaly Detection}
\label{sec:lifelong_anomaly_detection}

{

In this section, we start by {analyzing the challenges arising in dynamic real-world scenarios and} emphasizing the limitations of currently adopted anomaly detection approaches. Second, we discuss the advantages lifelong anomaly detection could bring to the anomaly detection landscapes. 
}

\subsection{When is non-lifelong anomaly detection not enough?}
Offline anomaly detection has shown to be useful in different applications where it is possible to gather and process background data, such as post-incident analysis \cite{goldstein2016comparative}, breast cancer detection \cite{9618843}
and gravitational waves detection \cite{corizzo2020scalable}. 
However, offline models are not sufficient for many real-world dynamic applications as they do not consider any change in the normal class.

Online anomaly detection methods 
partially address this limitation, providing learning systems with continuous updating capabilities. However, the underlying assumption is that only the most recent information is required to maintain satisfactory performance on the anomaly detection task. In this context, forgetting is a desirable property that allows the model to prevent obsolescence \cite{krawczyk2015one}. This behavior is considered to be sufficient to deal with many dynamic learning settings such as crowd anomaly detection \cite{FENG2017548} and fatigue detection \cite{laxhammar2013online}. 
It is also possible to detect whether concept drift occurred by monitoring the statistical properties or the model's error rate, and updating anomaly detection models to reflect the most recent conditions of the environment \cite{krawczyk2015one}. 
However, methods coupled with concept drift detection also follow the assumption that only the most recent data is relevant for the anomaly detection task. 
As a result, these methods are prone to forgetting past knowledge, which is a shortcoming in domains with recurring tasks. 

Many real-world domains may greatly benefit from the adoption of lifelong anomaly detection, as they are inherently characterized by dynamic and quickly evolving conditions, as well as recurring conditions. {These challenges require} model capabilities that foster simultaneous adaptation and knowledge retention.
%
In the following, we describe three out of many possible real-world domains where such model capabilities are required.

First, monitoring human conditions to detect harmful states must be able to deal with many human activities, each presenting a unique definition of the normal class. In this setting, new life habits bring new activities that can be assimilated as tasks to be learned by the model (e.g., jogging, characterized by a high heart rate that should not be considered anomalous behavior). 
Forgetting activities carried out in the past is not acceptable in this setting and brings a number of practical disadvantages, which are systematically described in Section \ref{sec:lifelong_anomaly_detection_drawbacks}.

Another example is the detection of intrusions in a cloud environment, which requires the ability to deal with a dynamic environment where multiple virtual servers (cloud instances) are added or removed over time. Such instances have different characteristics of normal behavior that depend on active services and user interactions. 
The system must be able to adjust and detect anomalies in traffic patterns in new cloud instances, but, at the same time, it should not decrease its performance when analyzing traffic from already monitored cloud instances. 

Looking at a different domain, the identification of faults and malfunctioning in cyber-physical systems, { such as water treatment plants or smart grids}, is characterized by a very dynamic environment with multiple operating conditions and different uncontrollable inputs (e.g., geophysical factors in nature) that change over time. Moreover, components also age over time or are replaced with other components with different specifications. In this context, the model should be able to deal with tasks corresponding to different operating conditions.

All these domains are characterized by {challenges such as} a number of evolving emerging conditions that require prompt model adaptation, as well as recurring conditions that require the ability to preserve the knowledge of previously observed conditions. This duality creates the ideal conditions for the adoption of lifelong anomaly detection. 
{However, to verify this assumption, it is important to assess whether lifelong scenarios impact the performance of non-lifelong anomaly detection models \textbf{(RQ1)}, and whether adopting knowledge retention capabilities actually results in an improvement for such models in lifelong scenarios \textbf{(RQ2)}.}

\subsection{Lifelong anomaly detection to the rescue}
\label{sec:lifelong_anomaly_detection_drawbacks}
\begin{figure}[]
\centering
\includegraphics[width=0.47\textwidth]{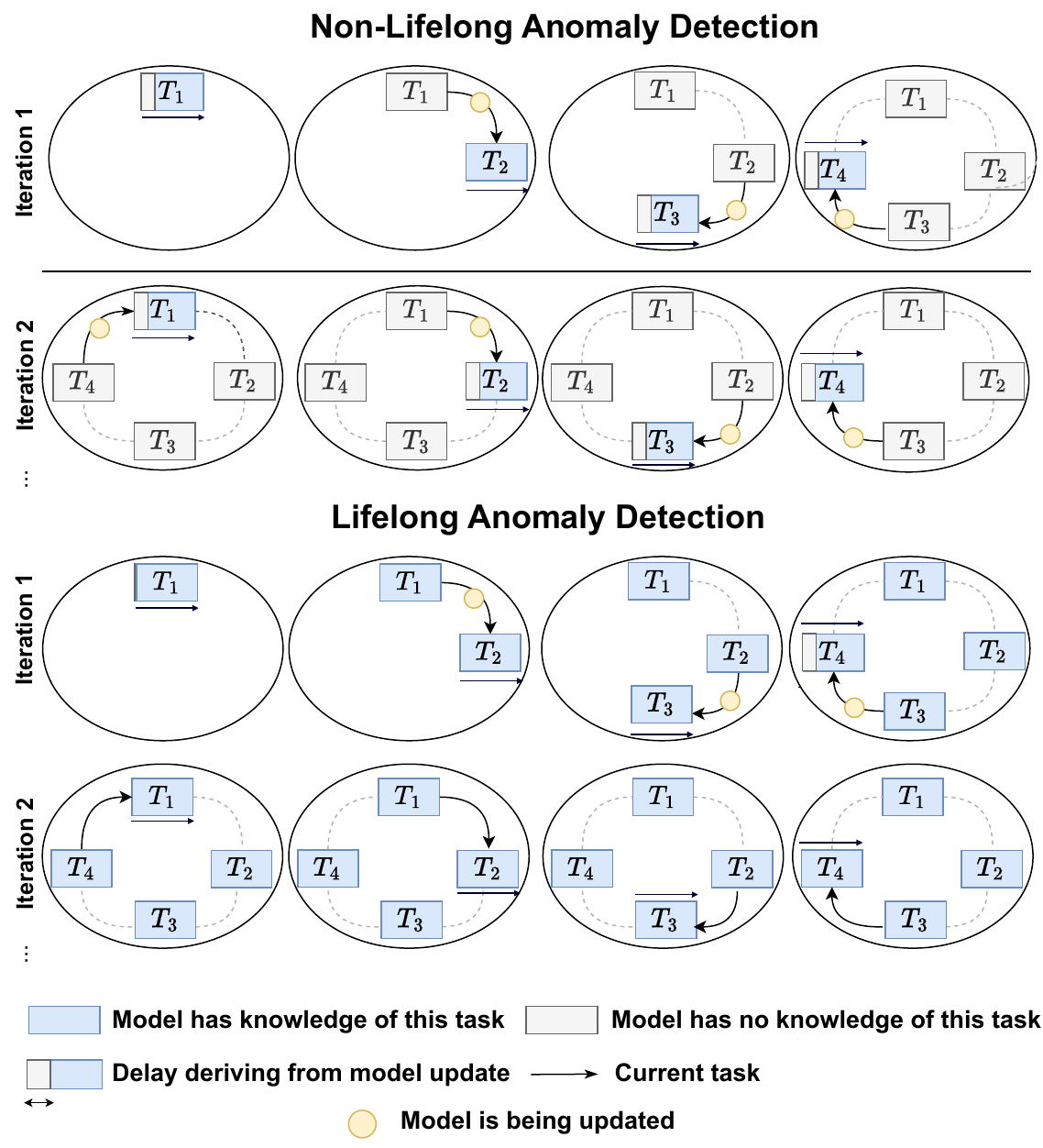}
\caption{A scenario with four recurring tasks $(T_1, T_2, T_3, T_4)$. Conventional anomaly detection requires constant model updates and results in detection delays. Lifelong learning mitigates this burden by retaining knowledge of tasks. 
}
\label{fig:recurrent_tasks}
\end{figure}

Figure \ref{fig:recurrent_tasks} shows a representative scenario that compares conventional anomaly detection with model updates to lifelong anomaly detection. In the second iteration, lifelong anomaly detection does not require model updates after a recurrence of each task. In contrast, conventional anomaly detection keeps updating the model, resulting in detection delays, i.e., false predictions, until the model has incorporated the new task. Moreover, a scenario with 100 iterations would require just 4 model updates for lifelong anomaly detection \textit{vs.} 400 model updates for conventional anomaly detection, during which detection delays will occur. 
Many real-world scenarios with recurrence could be mapped to it, including sequences of human activities, geophysical phenomena such as weather patterns, and operating conditions of cyber-physical systems.

\begin{figure*}[htp]
\centering
\includegraphics[width=0.90\textwidth]{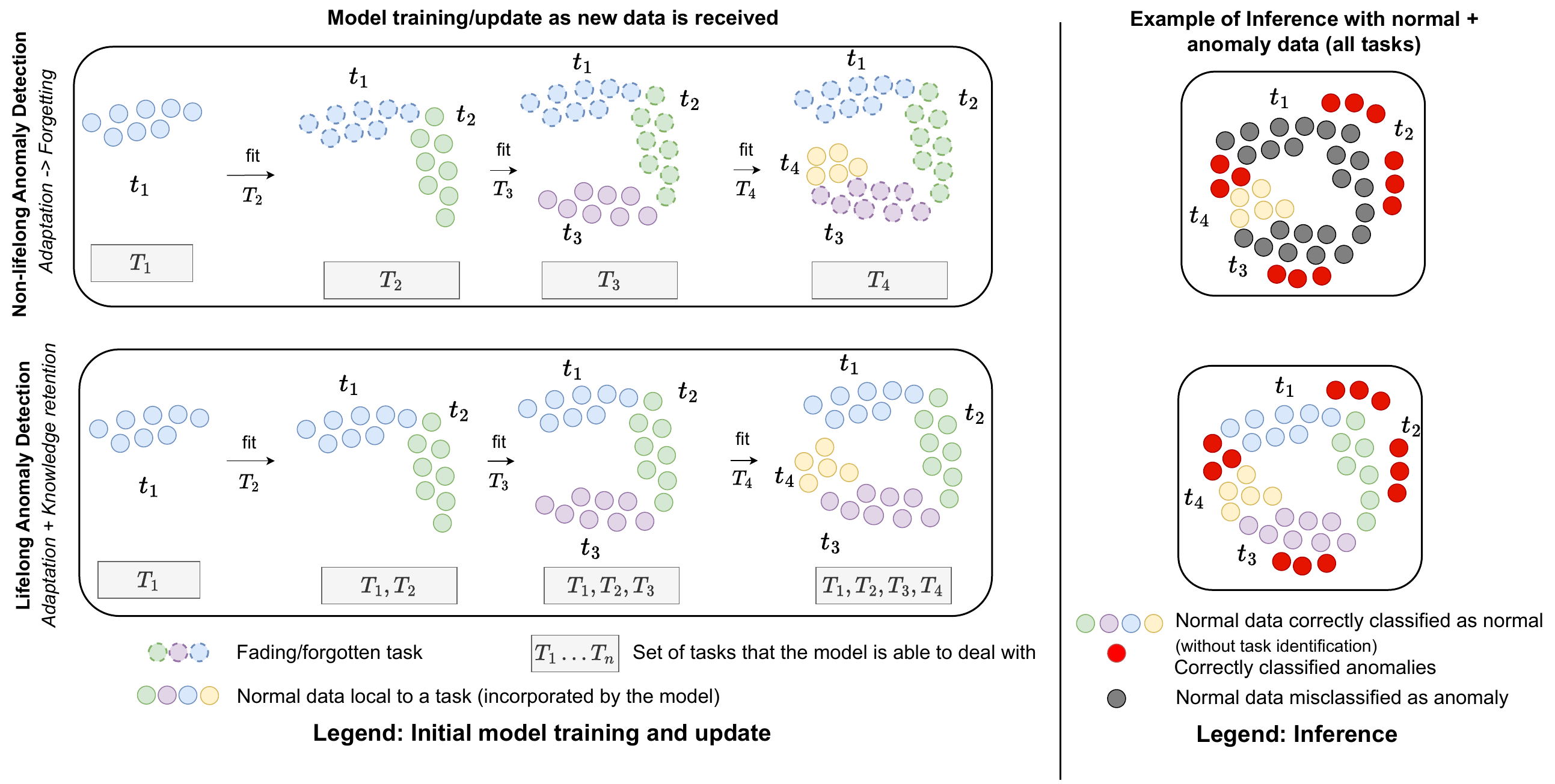}
\caption{Comparison of training/update and inference for non-lifelong and lifelong anomaly detection in the scenario with four tasks $(T_1, T_2, T_3, T_4)$. 
In non-lifelong anomaly detection, the model forgets the previous tasks as soon as a new task is learned (left -- top). In contrast, the lifelong anomaly detection model aims to retain  knowledge of all tasks (left -- bottom). 
This characteristic has a serious impact on the model's {behavior during inference} (right). In non-lifelong anomaly detection, after learning task $T_4$, the model misclassifies data from previous tasks as anomalous since it considers only data from the current task as normal behavior (right -- top). On the other hand, the ideal lifelong anomaly detection model retains the knowledge of all tasks, preventing the misclassification of normal data from previous tasks as anomalous (right -- bottom). 
{This difference in behavior between non-lifelong and lifelong  anomaly detection may lead to a discrepancy in their performance scores.}
}
\label{fig:method}
\end{figure*}

{

In Figure \ref{fig:method}, we show a comparison between the conventional anomaly detection approaches and a counterpart for anomaly detection that entails lifelong learning. In this example, normal class data evolves over time in a task-sequential manner. As new tasks are presented, the model is updated -- either in an online manner, possibly following concept drift detection, or in a lifelong learning manner.
It can be observed that the non-lifelong anomaly detection with a model update approach entails forgetting as a way to focus on the most recent data, leading to mistakenly classifying normal data from previous tasks as anomalous. 
On the other hand, the lifelong learning approach for model update aims at adapting the model to incorporate new tasks without forgetting previous tasks. 
The advantage is to exploit knowledge from the combination of the different tasks to provide a more comprehensive and robust anomaly detection model, which correctly identifies normal data from all tasks as normal behavior. 

Overall, even though the conventional anomaly detection with model update approach would be, in principle, able to re-learn previously forgotten tasks as they reoccur, this has several drawbacks, as indicated above. 
From a practical viewpoint, frequent model updates require additional computational resources in the presence of highly recurring tasks. 
Moreover, delays in the responsiveness of the model caused by the time difference between {the appearance} of a recurring task and updating the model once concept drift is detected can be costly in terms of false or missed detections. {We aim to verify this assumption throughout an experimental analysis involving non-lifelong anomaly detection models exposed to lifelong learning scenarios \textbf{(RQ1)}}.
}

By {analyzing the challenges pertaining to the lifelong anomaly detection learning setting and by generalizing} the examples presented in Figure \ref{fig:recurrent_tasks} and Figure \ref{fig:method}, we identify the following drawbacks of conventional anomaly detection with constant model updates:  
\begin{itemize}

\item By forgetting past knowledge and only adapting to the new normal class distribution, the system may trigger a  large number of false positives when recurring patterns are presented, leading to a dramatic decrease in performance until a retraining phase is undertaken;
\item Inability for the models to leverage skills learned in the past in combination with recent skills to solve tasks in a more compelling way;
\item Experimental settings that are too simplistic to reflect the complexity of many real-world challenges, which usually involve the appearance of new conditions and the recurrence of old conditions, requiring a more comprehensive evaluation across all tasks; 
\item A consistent use of computational resources for data processing and model training to deal with recurring tasks;
\item From a theoretical viewpoint, the model will not provide a comprehensive view of the environment without the possibility to leverage task similarity and knowledge transfer across a combination of tasks to solve every single task. 
   
\end{itemize}

These drawbacks make current anomaly detection models rather simplistic in comparison to sophisticated human-level intelligence {when faced with complexities and challenges brought by lifelong anomaly detection scenarios} {\textbf{(RQ1)}}. 
Intuitively, humans are exposed to different aspects of reality, building up skills incrementally throughout their lifespan and improving their general knowledge base.
Introducing similar capabilities in models should allow {them} to provide a more sophisticated behavior that translates into more reliable and accurate predictions \cite{kudithipudi2022biological} {\textbf{(RQ2)}}.

In practical terms, by leveraging discoveries
in biology, neuroscience, and computer science, lifelong learning enables the possibility of incorporating comprehensive knowledge in models. 
Examples include the adoption of task similarity, curriculum learning, yielding positive forward and backward transfer across different tasks, as already demonstrated in image classification \cite{chaudhry2018efficient}, 
object detection \cite{PARISI201954} and reinforcement learning problems \cite{abel2018policy}.

Overall, we argue that the adoption of lifelong learning in anomaly detection would yield the ability to consider the combination of all these aspects, providing more complex learning strategies that lead to more informed decisions. Instead of constantly forgetting and learning each individual condition, an ideal model could leverage past knowledge to retain performance across all conditions. {We aim to verify this assumption by designing experiments that uncover whether the adoption of lifelong learning knowledge retention strategies can be beneficial for non-lifelong anomaly detection models \textbf{(RQ2)}.}

In summary, the minimal set of advantages for the adoption  of a lifelong anomaly detection approach includes capabilities such as: \textit{i)} simultaneous adaptation and knowledge preservation; \textit{ii)} inference that exploits a more comprehensive knowledge of the domain or environment at hand; \textit{iii)} resource-savvy model updates compared to conventional anomaly detection methods with constant model updates; \textit{iv)} more realistic experimental settings and evaluation schemes that consider all tasks in combination.

\section{Lifelong Anomaly Detection: Scenarios and evaluation protocols}
\label{sec:scenarios}

Given the novelty of lifelong anomaly detection, 
as shown by the limited availability of research works on the subject, 
it is important to devise procedures and guidelines
to standardize its adoption. This section aims to {provide new perspectives on how to address lifelong learning challenges in anomaly detection. To this end, we} devise a categorization 
of lifelong learning scenarios, provide a scenario creation algorithm, and evaluation protocols that can guide researchers 
interested in the problem. 
\subsection{Lifelong Learning Scenarios: An Anomaly Detection Perspective}

Current lifelong learning approaches are focused on classification tasks, where \textit{tasks} are defined as sets of classes (e.g., in the MNIST dataset, any combination of two classes among its 10  classes), and the learning workflow encompasses a sequence of $n$ tasks $T = t_1, t_2, \dots, t_n$ where the model is challenged to learn new tasks without forgetting previous tasks.

Another important element of comparison is the type of \textit{learning scenarios}. {We recall that} lifelong image classification usually describes three types of scenarios: task-incremental, class-incremental, and domain-incremental \cite{van2019three}. 
{
These scenarios differ based on the availability of task labels and task boundaries. 
%
%
For all scenarios except domain-incremental, in the simple classification example mentioned above (MNIST), task labels identify a specific subset of 10 classes currently being presented to the model, whereas task boundaries identify the beginning/end of such task.
On the other hand, in a domain-incremental scenario, tasks represent a new distribution of already known classes, where task labels identify specific distributions and task boundaries indicate the moment when distribution changes. 
The availability of the task labels and boundaries depends on the degree of available domain knowledge \cite{sharma2018learning}.
A more detailed description of learning scenarios in lifelong classification may be found in \cite{van2019three}.
It is worth noting that emerging scenarios are being proposed to account for the limitations of previously existing ones. One example is the consideration of the temporal dimension in online lifelong learning scenarios \cite{ghunaim2023real}. 
}

The notion of a task in anomaly detection clearly differs from the conventionally adopted definition in image classification since we usually deal with two classes: normal and anomaly. 
Tasks in this context represent various aspects of the normal class, which is expected to evolve over time. 
Moreover, the normal class may also change its role depending on the context, i.e., what is normal in one context can be anomalous in another, further increasing the complexity of the problem. 
To differentiate lifelong anomaly detection setting, we define a self-consistent behavior\footnote{A behavior could correspond to a new distribution, change of a performed activity, or a new state of the environment, depending on the specific analytical context considered.} of the normal class, alongside the specific anomalies occurring with it, as a \textit{concept}.  

For instance, in monitoring human conditions to detect harmful states, the entire normal class can be thought of as a set of concepts: resting, jogging, and eating, all presenting different characteristics. 
In lifelong anomaly detection, multiple consistent behaviors of the normal class are presented over time instead of new classes as in class and task-incremental scenarios, and we are focused on the evolution of a single normal class instead of the evolution of all classes as in domain-incremental scenarios. 
{For example, a high heart rate can be considered an anomaly in resting conditions but is expected during jogging, and therefore it does not represent an anomaly in this context.}
Therefore, to deal with the inadequacy of lifelong image classification scenarios in the context of anomaly detection, we define distinctive scenarios for this setting.
Following the example of anomaly detection in human conditions, 
\textit{concept identifiers} define a consistent behavior of the normal class (a specific activity), whereas  \textit{concept boundaries} represent explicit information on whether the currently analyzed concept (a specific activity) has changed. Concept identifiers and concept boundaries may correspond to task labels and task boundaries, respectively, in lifelong image classification. 

Based on this consideration, we identify the following learning scenarios in increasing order of complexity: 
\begin{itemize}
    \item{\textit{Concept-aware}: Known concept identifier and concept boundaries. }
    \item{\textit{Concept-incremental}: Unknown concept identifier but known concept boundaries}.
    \item{\textit{Concept-agnostic}: Unknown concept identifier and concept boundaries}
\end{itemize}

In reference to our example of anomaly detection in human conditions, a concept-aware scenario implies that the model is aware of the currently processed activity and its lifespan (at both training and inference time). On the other hand, a concept-incremental scenario only provides an indication that a change of activity has occurred without any identifying information about the specific activities. Finally, a concept-agnostic scenario is the most challenging, as it does not provide any supporting information about the current activity being performed and its lifespan. 
These notions are general and can be adopted in any domain.

\subsection{Scenarios design}
\label{section:scenarios}

In the following, we propose a scenario design procedure that applies to most datasets and enables researchers and practitioners to transition their current scenarios and evaluation setup toward lifelong anomaly detection.


Algorithm \ref{alg:scenario-design} presents a general pseudo-code for scenario design. 
{It requires users to define a few parameters, which determine the creation of diverse scenarios: the number of desired concepts $\mathbf{c}$, Normal ($\mathbf{N}$), and Anomaly ($\mathbf{A}$) data from a given dataset, and three functions: $\phi$, $\gamma$, and $\lambda$.
The algorithm leverages concept creation functions $\phi$ and $\gamma$ to create normal and anomaly concepts based on normal and anomaly data, respectively. Their goal is to transform the original dataset into self-consistent sets of data points having common characteristics.
An example implementation for $\phi$ and $\gamma$ is a clustering algorithm of choice, based on user preferences. 
%
The assignment function $\lambda$ matches each normal concept with an anomaly concept, leading to a combined concept containing one normal and one anomaly concept, which allows for model training and evaluation. The sequence of these combined concepts defines the complete scenario. An example implementation for $\lambda$ is mapping an anomaly cluster to its closest normal cluster. 
}

{Focusing on Algorithm} \ref{alg:scenario-design}, first, we create concepts for the normal class through a concept creation function $\phi$ (Line 1). 
{
The concept creation function can leverage any aspect or feature value that allows us to delineate the boundaries of one concept.
}
Second, we create concepts for the anomaly class through anomalous concepts creation function $\gamma$ (Line 2). Third, 
for each normal concept $C_{N_i}$, we select a corresponding anomaly concept $C_{A_j}$ using a function $\lambda$
(Line 3-5). 
The combination of $C_{N_i}$ and $C_{A_j}$ is a concept added to the lifelong scenario (Line 6). 
Each time a concept is built, the selected anomaly concept $C_{A_j}$ is removed from the set of available anomaly concepts $C_{A}$ {so that it appears only once throughout the scenario}  (Line 7).
The algorithm returns the resulting scenario as a sequence of concepts (Line 9), each of which may need to be separated into training and evaluation data depending on the learning settings, e.g., unsupervised or semi-supervised.

    \begin{algorithm}
    \small
    \SetAlgoLined
    \DontPrintSemicolon
    \LinesNumbered
    \SetKwInput{Input}{Input}
    \SetKwInOut{Parameter}{Parameters}
    
    \Input{$c$ -- Number of desired concepts} 
    \Input{$\mathbf{N}, \mathbf{A}$ -- Normal/Anomaly data} 
    \Input{$\phi$ -- Concepts creation function for normal data}
    \Input{$\gamma$ -- Concepts creation function for anomalies} 

    \Input{$\lambda$ -- Assignment function} 
    $C_N \gets \phi(\textbf{{N}},c)$ \tcp*{{Create concepts} $\{ C_{N_0}, C_{N_1}, \dots, C_{N_c} \}$} 
    $C_A \gets \gamma(\textbf{{A}},c)$ \tcp*{{Create concepts} $\{ C_{A_0}, C_{A_1}, \dots, C_{A_c} \}$} 
    $T \gets \emptyset$ \tcp*{{Result scenario}}
    \For{$C_{N_i} \in C_N$}{
        $j \gets \lambda(C_A, C_{N_i})$ \tcp*{{Match anomaly-normal concepts}}
        $T \gets T \cup (C_{N_i}, C_{A_j})$ \tcp*{{Add concepts to  scenario}}
        $C_A \gets C_A - C_{A_j}$ \tcp*{{Remove used anomaly concept}}
    }
    \Return $T$ \\
    \caption{Scenario design protocol}
    \label{alg:scenario-design}
    \end{algorithm}


\begin{figure}[]
\centering
\includegraphics[width=0.5\textwidth]{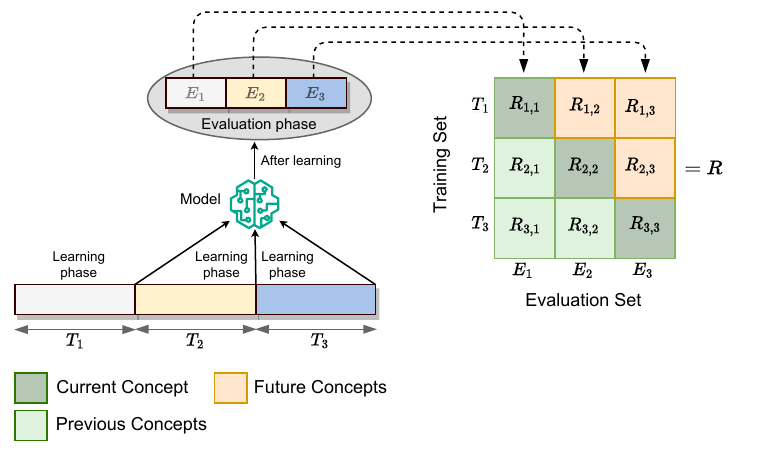}
\caption{Lifelong evaluation protocol. {The model handles a sequence of concepts $i = {1, 2, 3}$. For each concept $i$, the model is trained on  training set $T_i$ (learning phase). After each learning phase, the evaluation phase is triggered, where the model anomaly detection performance (in terms of ROC-AUC) is evaluated on all testing sets $E_j$ from all concepts (previous, current, and future)}. The evaluation protocol creates a matrix $R$, in which the entry $R_{i, j}$ represents model performance {in terms of ROC-AUC} on concept $j$ after learning concept $i$. This matrix is used to compute final metric values, such as Lifelong ROC-AUC, BWT, and FWT.
}
\label{fig:eval-protocol}
\end{figure}

The proposed method allows us to create diverse scenarios depending on the user selection of $\phi$, $\gamma$, and $\lambda$. 
%
For example, it is possible to leverage k-Means to cluster normal and anomaly concepts (leveraging $\phi$, and $\gamma$ functions) and assign each anomaly concept to the closest normal concept (leveraging $\lambda$ function).
Alternative scenarios can be designed by customizing $\phi$, $\gamma$, and $\lambda$, paving the way for a wide range of possible scenarios.
To simplify this task, our code is publicly available and can be used to easily generate scenarios for any dataset chosen by users: 
\url{https://github.com/lifelonglab/lifelong-anomaly-detection-scenarios}.
{We note that different choices of $\phi$ and $\gamma$ may lead to concept imbalance, i.e., some concepts may present significantly fewer samples than others. This situation may be problematic in some settings, or exacerbate the learning complexity for some anomaly detection models. In these cases, consideration of strategies for imbalanced learning such as resampling {\cite{8737955}}, cost-sensitive learning, and special-purpose algorithms {\cite{ghosh2022class}}. 
It is worth noting that, in our experiments, we did not experience high imbalance ratios for concepts generated with our protocol.}

\subsection{Model evaluation}
\label{sec:evaluation}

Lifelong learning scenarios require a continuous evaluation across all concepts. 
To realize this goal, we adopt a lifelong learning evaluation protocol that considers the performance of all concepts across a learning scenario. 

Algorithm \ref{alg:experiments}  provides a general overview applicable to a vast number of use cases. Without loss of generality, the protocol can be modified to accommodate specific requirements, e.g., recurring concepts, time-based concepts, unsupervised learning, etc.
{Moreover, our protocol can support any base model of choice, and any data preprocessing step, such as data augmentation and missing data treatment, to deal with specific data challenges.}
{

First, the evaluation protocol initializes a matrix $R$ to accommodate anomaly detection results for specific tasks (Line 1). Second, the protocol iterates over training sets for all concepts (Line 2). For each concept, the model is trained/updated (Line 3) and evaluated on all testing sets for all concepts (Lines 4-{5}), i.e.{,} previous, current, and future concepts. Our protocol yields a matrix  $R$, where entries $R_{i, j}$ define the ROC-AUC metric of the model evaluated on concept $j$ after learning concept $i$. 
A graphical representation of this protocol is shown in Figure \ref{fig:eval-protocol}.

}

    \begin{algorithm}
    \small
    \SetAlgoLined
    \DontPrintSemicolon
    \LinesNumbered
    \SetKwInput{Input}{Input}
    \SetKwInOut{Parameter}{Parameters}
    \Input{$\mathbf{T}$ -- {S}equence of $N$ training sets} 
    \Input{$\mathbf{E}$ -- {S}equence of $N$ testing sets} 
    \Input{$L$ -- {L}earning model} 
    \Input{$\rho$ -- {Evaluation function computing ROC--AUC}}

    $R_{N \times N} = \{ \}$ \tcp*{{Initialize results matrix $NxN$}}
    \For{$T_i \in \mathbf{T}$}{
        {$L \gets $ update $L$ with $T_i$} \tcp*{{Train/update model}} 
        
        \For{$E_j \in \textbf{E}$}{
            {$R_{i, j} \gets \rho(L, E_j)$} \tcp*{{Evaluate $L$ on $E_j$}}
        }
    }
    \Return $R$ \\
    \caption{Pseudo-code of the evaluation protocol}
    \label{alg:experiments}
    \end{algorithm}

The matrix $R$ can be used to directly compute lifelong learning metrics, such as backward and forward transfer. 
{

These metrics allow us to assess model behavior more extensively than standard performance metrics by taking into account the model's performance on different concepts (previous, current, and future).  
}
Inspired by \cite{diazrodriguez2018dont}, we propose a \textbf{Lifelong ROC-AUC} -- a lifelong variant of ROC-AUC that can adequately assess models' performance on all concepts after learning every new concept, instead of models' performance on just a single concept. 
It is defined as:
\begin{equation}
    \text{Lifelong ROC-AUC} = \frac{\sum_{i \ge j}^N R_{{i,j}}}{\frac{N(N+1)}{2}}
    \label{eq:roc-auc}
\end{equation}
{

The metric is computed considering previously learned concepts, including the current concept, which corresponds to averaging over $\frac{N(N+1)}{2}$ entries from lower triangular. 
}
We favor ROC-AUC over threshold-dependent metrics such as Precision, Recall, and F--Score, since it allows us to evaluate the model's performance more comprehensively. ROC--AUC may be swapped with other metrics of choice without impacting the validity of the protocol. 

\textbf{Backward Transfer for ROC--AUC ($BWT$)} measures the impact of learning new concepts on the performance of all previously learned concepts. Negative backward transfer suggests that the model is prone to forgetting. A strongly negative value is also sometimes regarded as catastrophic forgetting. On the other hand, positive backward transfer suggests that learning new concepts benefits models' performance on previously learned concepts.  Backward transfer is computed over all concepts as:
    \begin{equation}
        BWT = \frac{\sum_{i=2}^N\sum_{j=1}^{i-1} R_{i, j} - R_{j,j}}{\frac{N(N-1)}{2}}
        \label{eq:bwt_all}
    \end{equation}
    
The impact of learning each concept on the model's performance on future concepts is measured by \textbf{Forward Transfer for ROC--AUC ($FWT$)}. 
Forward transfer can {also be} thought of as the zero-shot model performance on future concepts since it assesses model performance on unseen concepts. It partially depends on concept similarity (task similarity) and {the} model's knowledge transfer ability. It is computed as:
    \begin{equation}
        FWT = \frac{\sum_{i<j}^{N} R_{i, j}}{\frac{N(N-1)}{2}}
        \label{eq:fwt_all}
    \end{equation}

It is noteworthy that the protocol slightly differs based on the learning setting. Specifically, in concept-aware and concept-incremental scenarios, batches $T_i$ (training) and $E_i$ (evaluation) correspond to the single $i-$th concept. As for concept-agnostic settings, a batch does not necessarily correspond to a single concept since the setting assumes that no explicit concept boundaries are provided to the lifelong algorithm.
As a result, the evaluation may require considering multiple batches as belonging to the same concept or a single batch including data for more than one concept.

\section{Experiments}
\label{sec:experiments}
\begin{figure*}[h!]
    \centering
    \includegraphics[width=1.0\textwidth]{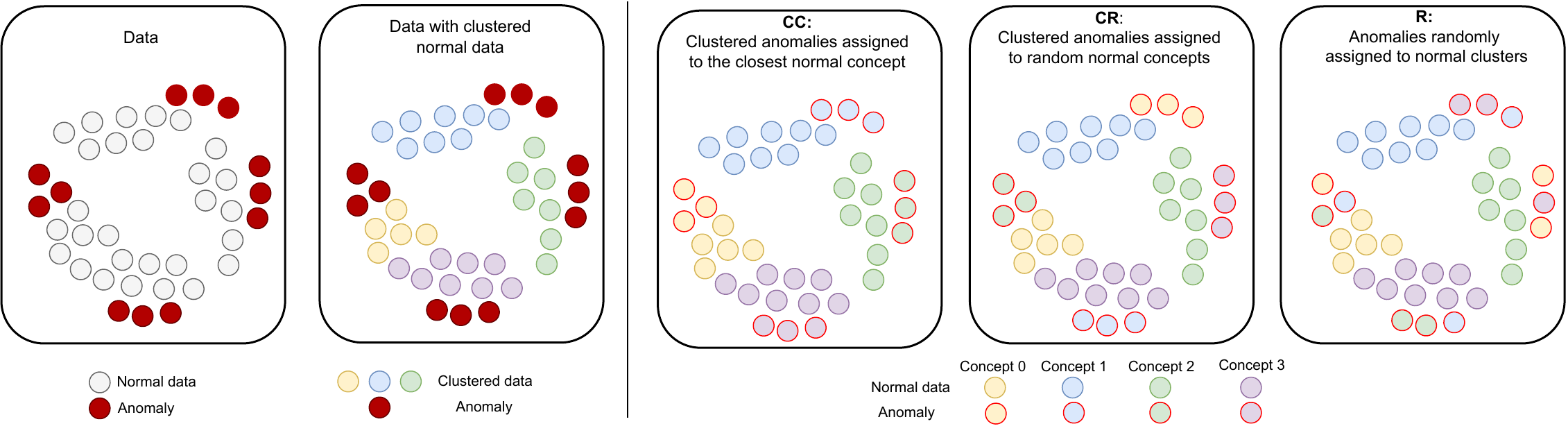}
    \caption{Lifelong scenarios variants based on different choices of concept creation functions $\gamma$ and $\lambda$: \textit{i)} clustered anomaly concepts assigned to the closest normal concept (\textbf{CC}), \textit{ii)} clustered anomaly concepts assigned randomly to normal concepts (\textbf{CR}), and \textit{iii)} anomalies randomly assigned to normal concepts (\textbf{R}). }
    \label{fig:conceptual-scenarios}
\end{figure*}

{
 
Our experiments are directed at answering two main research questions: \textit{i)} Do lifelong scenarios impact the performance of non-lifelong anomaly detection models?; ii) Does adopting lifelong learning provide a valuable improvement in the learning capabilities of anomaly detection models in lifelong scenarios? We empirically address these two questions in the following two subsections {and provide insights on anomaly detection performance, forgetting, and task similarity}. 
}
%
{To answer the above question, we design three lifelong scenario variants based on different choices:} 
\begin{itemize}
 \item  \textbf{CC}: {clustered anomaly concepts assigned to the closest normal concept, where $\gamma$ and $\phi$ leverage a clustering function, and $\lambda$ maps a given normal concept to the closest anomaly concept.}
 \item  \textbf{CR}: {clustered anomaly concepts assigned randomly to normal concepts, where $\gamma$ and $\phi$ leverage a clustering function, and $\lambda$ assigns a random anomaly concept to a given normal concept.}
 \item \textbf{R}: {anomalies randomly assigned to normal concepts, where $\gamma$ leverage a clustering function, $\phi$ creates anomaly concepts using random sampling, and $\lambda$ assigns a random anomaly concept to a given normal concept.}
\end{itemize}
{For all scenario variants, we use k-Means as the clustering function.}
{

{
These scenarios are conceptually represented as two-dimensional plots in Figure \ref{fig:conceptual-scenarios}.}
{In our experiments, we extract between $5$ and $20$ concepts depending on the complexity and the size of each dataset.}

}

{
In our experiments, we employ a diverse set of datasets encompassing cybersecurity and smart grids.}
\begin{itemize}

\item{{\textbf{NSL-KDD} {\cite{nslkdd}}: Widely adopted dataset containing records of network traffic gathered during the DARPA Intrusion Detection Systems evaluation program;}}
\item{{\textbf{UNSW-NB15} {\cite{unsw}}: The dataset containing records of network traffic describing hybrid real modern normal and  contemporary synthesized attack activities;}}
\item{{\textbf{Energy} {\cite{corizzo2019anomaly}}: Sensor-based anomaly detection in photovoltaic/solar power plants. The data was collected from power plants located in Italy (17 plants, 2.5 years);}} 
\item{{\textbf{Wind} {\cite{corizzo2021spatially}}: Wind power production dataset, containing anomalous patterns occurred in eolic/wind parks (Wind). The data was modeled using the Weather Research \& Forecasting (WRF) model (5 plants, 2 years).}}
\end{itemize}
{All datasets present fairly different feature representations and represent data collected by heterogeneous systems. They also present different types of anomalies and complexity levels. 
}

We use five popular anomaly detection models described in Section \ref{sec:background}: 
One-Class Support Vector Machines (OC-SVM), Local Outlier Factor (LOF), Isolation Forests (IF), Variational Auto-Encoders (VAE), and Copula-based Outlier Detection (COPOD).

{We leverage the following learning strategies:}
\begin{itemize}\item{\textbf{Naive lifelong}: models are updated as new data becomes available, without any smart lifelong learning strategy to tune adaptation and knowledge retention. {By updating the model only based on the new data, a reasonable expectation is that the model will gradually or catastrophically forget knowledge of previously presented data. It can be considered as a lower-bound non-lifelong baseline learning strategy}.}
\item{\textbf{Multiple Single-Task Experts (MSTE)}: a way to simulate upper-bound model performance in a non-lifelong scenario. {In this strategy, a pool or ensemble of models, each of which is an expert for a single concept, is adopted. Whenever a new concept is presented, a new model is trained on the new data and added to the pool}. We note that this is an unrealistic setting since, in real-world scenarios, it requires extremely high computational resources to deal with a potentially infinite number of models, and the availability of concept identifiers, which are not available for concept-incremental and concept-agnostic scenarios.}
\item{{\textbf{Replay}}}: a replay-based method that preserves selected data samples from previous concepts in a memory buffer, which is limited in size by a parameter known as a budget. When the model faces a new task (concept), the replay buffer is updated to include the data from the new concept. {As a result, the replay buffer contains knowledge of all concepts presented so far. The replay buffer is then used while updating the model to mitigate forgetting {\cite{buzzega2021rethinking}}.
The expectation is that, by providing a summarized representation of all concepts, the model should, in principle, be able to preserve a satisfactory performance on all concepts, without a significant degree of forgetting for any of the concepts}.  In our experiments, we use a simple balanced replay buffer with a very constrained budget of 3,000 data samples.
\end{itemize}

{To this end, we adopt the learning evaluation workflow in Algorithm \mbox{\ref{alg:experiments}}, which provides the model with the challenge of adapting to new data while retaining knowledge of previously observed data. Technically, we create a matrix where the performance of the model is measured on all concepts after learning each single concept. Metric computation takes place according to Equations \mbox{\ref{eq:roc-auc}}-\mbox{\ref{eq:fwt_all}}.}

{We address \textbf{RQ1} by devising experiments to assess whether} non-lifelong anomaly detection methods are impacted by the challenges brought by lifelong scenarios. 
To this aim, leverage the two mentioned strategies (Naive lifelong and MSTE) to showcase if there is a gap between models' performance in non-lifelong scenarios vs. lifelong scenarios, which would suggest the need for adopting lifelong learning strategies. 
{Particularly, we are interested in observing whether MSTE achieves higher performance values in terms of ROC-AUC, BWT, and FWT compared to Naive. {This expectation is motivated by the observation that Naive only adapts to new data without any knowledge retention mechanism, while MSTE does not experience forgetting due to the creation of a single expert model for each task}.}

{We address \textbf{RQ2} by devising experiments to assess whether the adoption of lifelong learning strategies has the potential to increase the performance of non-lifelong anomaly detection models in lifelong scenarios.}
{To this aim}, we analyze the impact of the adoption of a lifelong Replay strategy on the performance of all base models {to show that the adoption of lifelong learning strategies may be beneficial to improve model performance in a lifelong anomaly detection scenario}.
{Experimental results\footnote{We note that our results are obtained by averaging multiple executions with different hyperparameter values (see Appendix \ref{sec:hyperparameters}) to showcase the reliability of the results, similarly to a cross-validation evaluation {\cite{khan2022higher, khan2021improved}}.} are shown in Tables} \ref{tab:results}, \ref{tab:results-replay}.

{
\begin{table*}[h!]
    \centering
    \footnotesize
    \setlength{\tabcolsep}{2pt}
    \caption{Experimental results for Naive lifelong strategy with all methods (IF, LOF, COPOD, OC-SVM, and VAE) and datasets (Energy, NSL-KDD, UNSW, Wind) in the three concept-incremental scenarios (CC, CR, R), according to ROC-AUC, BWT,  FWT metrics. For ROC-AUC, we also report results obtained with multiple single-task experts (MSTE) in parenthesis (upper bound).}
    \begin{tabular}{lccccccccccccccc} \\ 
\toprule
& \multicolumn{3}{c}{IF} & \multicolumn{3}{c}{LOF} & \multicolumn{3}{c}{COPOD} & \multicolumn{3}{c}{OC-SVM} & \multicolumn{3}{c}{VAE} \\ 
\textbf{Dataset} & ROC-AUC & BWT & FWT & ROC-AUC & BWT & FWT & ROC-AUC & BWT & FWT & ROC-AUC & BWT & FWT & ROC-AUC & BWT & FWT \\ 
\midrule
 Energy (CC) & 0.64 (0.88) & -0.29 & 0.61 & 0.71 (0.96) & -0.31 & 0.81 & 0.85 (0.91) & -0.07 & 0.90 & 0.76 (0.94) & -0.22 & 0.84 & 0.75 (0.92) & -0.21 & 0.81 \\ 
 Energy (CR) & 0.65 (0.97) & -0.40 & 0.53 & 0.70 (0.99) & -0.35 & 0.73 & 0.91 (0.98) & -0.09 & 0.86 & 0.75 (0.99) & -0.29 & 0.71 &  0.74 (0.99) & -0.31 & 0.68 \\ 
 Energy (R) & 0.59 (0.97) & -0.46 & 0.56 & 0.66 (1.00) & -0.41 & 0.70 & 0.89 (0.98) & -0.11 & 0.87 & 0.69 (0.99) & -0.37 & 0.70 & 0.68 (0.99) & -0.38 & 0.69 \\ 
 NSL-KDD (CC) & 0.68 (0.97) & -0.32 & 0.77 & 0.62 (0.93) & -0.33 & 0.56 & 0.54 (0.66) & -0.14 & 0.43 & 0.67 (0.99) & -0.36 & 0.85 & 0.67 (0.98) & -0.35 & 0.85 \\ 
NSL-KDD (CR) & 0.70 (0.96) & -0.28 & 0.72 & 0.60 (0.94) & -0.37 & 0.54 & 0.52 (0.62) & -0.13 & 0.44 & 0.75 (0.96) & -0.23 & 0.73 & 0.73 (0.96) & -0.25 & 0.73 \\ 
 NSL-KDD (R) & 0.85 (0.99) & -0.16 & 0.82 & 0.63 (0.95) & -0.35 & 0.52 & 0.74 (0.81) & -0.08 & 0.67 & 0.88 (1.00) & -0.13 & 0.85 & 0.87 (1.00) & -0.14 & 0.86 \\ 
UNSW (CC) & 0.49 (0.73) & -0.29 & 0.45 & 0.57 (0.88) & -0.38 & 0.45 & 0.32 (0.43) & -0.11 & 0.29 & 0.59 (0.78) & -0.22 & 0.53 & 0.55 (0.81) & -0.32 & 0.50 \\ 
 UNSW (CR) & 0.60 (0.94) & -0.41 & 0.50 & 0.67 (0.98) & -0.38 & 0.49 & 0.48 (0.67) & -0.24 & 0.24 & 0.73 (0.97) & -0.31 & 0.56 & 0.72 (0.98) & -0.32 & 0.51 \\ 
 UNSW (R) & 0.53 (0.90) & -0.45 & 0.45 & 0.60 (0.96) & -0.42 & 0.46 & 0.60 (0.76) & -0.18 & 0.44 & 0.55 (0.91) & -0.44 & 0.46 & 0.56 (0.94) & -0.46 & 0.44 \\ 
 Wind (CC) & 0.83 (0.90) & -0.10 & 0.74 & 0.65 (0.98) & -0.49 & 0.58 & 0.89 (0.93) & -0.06 & 0.89 & 0.77 (0.96) & -0.28 & 0.77 & 0.76 (0.96) & -0.30 & 0.76 \\ 
 Wind (CR) & 0.74 (0.95) & -0.32 & 0.71 & 0.62 (0.99) & -0.57 & 0.53 & 0.89 (0.96) & -0.12 & 0.90 & 0.72 (1.00) & -0.42 & 0.70 & 0.68 (0.98) & -0.46 & 0.73 \\ 
 Wind (R) & 0.74 (0.95) & -0.31 & 0.69 & 0.65 (0.99) & -0.52 & 0.50 & 0.90 (0.97) & -0.11 & 0.91 & 0.74 (0.99) & -0.38 & 0.66 & 0.71 (0.99) & -0.42 & 0.64 \\
\bottomrule
\end{tabular} \\

    \label{tab:results}
\end{table*}

\begin{table*}[h!]
    \centering
    \footnotesize
    \setlength{\tabcolsep}{2pt}
    \caption{Experimental results for Replay strategy with all methods (IF, LOF, COPOD, OC-SVM, and VAE) and datasets (Energy, NSL-KDD, UNSW, Wind) in the three concept-incremental scenarios (CC, CR, R), according to ROC-AUC, BWT,  FWT metrics.}    \begin{tabular}{lrrrrrrrrrrrrrrr} \\ 
\toprule
 & \multicolumn{3}{c}{IF} & \multicolumn{3}{c}{LOF} & \multicolumn{3}{c}{COPOD} & \multicolumn{3}{c}{OC-SVM} & \multicolumn{3}{c}{VAE} \\ 
& ROC-AUC & BWT & FWT & ROC-AUC & BWT & FWT & ROC-AUC & BWT & FWT & ROC-AUC & BWT & FWT & ROC-AUC & BWT & FWT \\ 
\midrule
 Energy (CC) & 0.82 & -0.06 & 0.64 & 0.95 & -0.01 & 0.78 & 0.89 & -0.01 & 0.89 & 0.83 & -0.04 & 0.85 & 0.77 & -0.16 & 0.83 \\ 
 Energy (CR) & 0.78 & -0.13 & 0.51 & 0.96 & -0.01 & 0.64 & 0.89 & -0.04 & 0.83 & 0.85 & -0.06 & 0.77 & 0.81 & -0.15 & 0.66 \\ 
 Energy (R) & 0.75 & -0.18 & 0.58 & 0.97 & -0.02 & 0.61 & 0.87 & -0.06 & 0.83 & 0.84 & -0.07 & 0.76 & 0.75 & -0.25 & 0.68 \\ 
 NSL-KDD (CC) & 0.85 & -0.11 & 0.92 & 0.89 & -0.03 & 0.33 & 0.79 & 0.00 & 0.78 & 0.71 & -0.21 & 0.89 & 0.82 & -0.09 & 0.93 \\ 
 NSL-KDD (CR) & 0.81 & -0.07 & 0.82 & 0.88 & -0.02 & 0.39 & 0.72 & -0.02 & 0.62 & 0.79 & -0.11 & 0.74 & 0.84 & -0.04 & 0.76 \\ 
 NSL-KDD (R) & 0.95 & -0.03 & 0.90 & 0.90 & -0.01 & 0.34 & 0.92 & 0.04 & 0.84 & 0.93 & -0.03 & 0.86 & 0.93 & -0.01 & 0.85 \\ 
 UNSW (CC) & 0.51 & -0.03 & 0.35 & 0.83 & -0.00 & 0.32 & 0.43 & 0.00 & 0.29 & 0.56 & -0.02 & 0.54 & 0.65 & -0.12 & 0.38 \\ 
 UNSW (CR) & 0.68 & -0.02 & 0.45 & 0.89 & -0.01 & 0.39 & 0.67 & 0.00 & 0.30 & 0.82 & 0.01 & 0.57 & 0.73 & -0.05 & 0.33 \\ 
 UNSW (R) & 0.48 & -0.08 & 0.35 & 0.82 & -0.05 & 0.40 & 0.52 & -0.05 & 0.39 & 0.50 & -0.08 & 0.46 & 0.66 & -0.08 & 0.39 \\ 
 Wind (CC) & 0.91 & -0.01 & 0.74 & 0.97 & -0.01 & 0.60 & 0.92 & -0.01 & 0.90 & 0.91 & -0.04 & 0.83 & 0.86 & -0.13 & 0.72 \\
 Wind (CR) & 0.88 & -0.08 & 0.70 & 0.97 & -0.03 & 0.57 & 0.89 & -0.06 & 0.89 & 0.88 & -0.08 & 0.75 & 0.84 & -0.18 & 0.74 \\ 
 Wind (R) & 0.90 & -0.06 & 0.67 & 0.98 & -0.02 & 0.52 & 0.91 & -0.06 & 0.89 & 0.91 & -0.06 & 0.72 & 0.86 & -0.17 & 0.72 \\ 
\bottomrule
\end{tabular} \\ 

    \label{tab:results-replay}
\end{table*}

\section{Results discussion}
\label{sec:discussion}
{In this section, we discuss results alongside two main perspectives: the impact of lifelong scenarios on non-lifelong anomaly detection problems, and the impact of the adoption of lifelong learning strategies providing knowledge retention capabilities.}

\subsection{Impact of lifelong scenarios on non-lifelong anomaly detection}

In this subsection, we focus on {providing insights on} how non-lifelong anomaly detection methods are impacted by the challenges brought by lifelong scenarios {(\textbf{RQ1})}. 
%




}
We present the experimental results in Table \ref{tab:results}.
%
The general trend that can be observed across all methods and datasets is that there is a gap between the anomaly detection performance in terms of ROC-AUC achieved by widely adopted anomaly detection methods and the hypothetical upper-bound defined by multiple single-task experts (MSTE). Notably, with the Energy dataset, base models with a Naive learning strategy are significantly outperformed by the MSTE approach (for example, for Isolation Forest, CC: $0.64$ vs. $0.88$ --- CR: $0.65$ vs. $0.97$ --- R: $0.59$ vs. $0.97$). This is not an isolated case but applies to the other datasets as well.
{

It is noteworthy that MSTE achieves very high performance in most cases, highlighting that the scenario generation procedure decomposes each dataset's complexities into sub-complexities (concepts), which are much more manageable to learn in isolation but much more challenging when provided as a lifelong scenario. 
This phenomenon has been observed in \cite{sharma2018learning} in the context of non-lifelong one-class classification.
Overall, it looks clear that non-lifelong anomaly detection methods are penalized in lifelong scenarios, leading to sub-optimal performance scores{, as evident by the low ROC-AUC values}.
{We present a graphical illustration of this phenomenon in Figure {\ref{fig:results-plot}}, which shows a clear performance gap between non-lifelong and lifelong strategies.}

Another lifelong metric worth analyzing is the backward transfer (BWT) {as it allows us to analyze how learning new tasks affects model performance on previous tasks}. {We recall that negative values of BWT indicate that learning new tasks introduces forgetting in previously learned tasks, whereas positive values are indicative of effective knowledge transfer capabilities across tasks (see Section \mbox{\ref{sec:evaluation}}).} We observe that all base models with Naive strategy present negative values of BWT, e.g., with Wind (R), models showcase values from $-0.11$ to $-0.52$, which shows that they are affected by a degree of forgetting (from mild to catastrophic).
{
This phenomenon is more clearly visible in Figure \ref{fig:wind_vae} (VAE) and \ref{fig:wind_LOF} (LOF), where the performance on $C_0$ drops from 1 to $0.69$ (VAE) and from $0.99$ to $0.41$ (LOF) after the last concept $C_4$ is learned. 
Negative backward transfer is also observed for $C_2$, where model performance gradually drops from $0.99$ to $0.5$ (VAE) and from $0.99$ to $0.34$ (LOF) after learning $C_3$. Subsequently, learning $C_4$ leads to increased forgetting of $C_2$, and the performance on $C_2$ drops to $0.078$ (VAE) and $0.098$ (LOF).
Another exciting aspect that can be observed is a positive backward transfer, which indicates that learning a new concept improves the performance of the model on a previous concept. This is emphasized by the performance on $C_0$ before and after learning $C_2$. Results show that learning $C_2$ increases the performance on $C_0$ from $0.51$ to $0.96$ (VAE) and from $0.42$ to $0.55$ (LOF). It means that the model can leverage the knowledge acquired while learning $C_2$ to present better anomaly detection capabilities on concept $C_0$ {by leveraging similarity between concepts (task-similarity)}.
{By analyzing BWT, we uncover concept similarity relationships and measure models' ability to retain and reuse knowledge for improving its overall performance across all concepts.}
}

Finally, moving our focus to Forward Transfer ($FWT$), we can observe values from $0.24$ -- UNSW (CR) with COPOD to $0.91$ -- Wind (R) with COPOD. This result suggests a moderate-to-high concept similarity that could be leveraged by models. Interestingly, these values are higher than those commonly seen in image classification since the one-class learning setting is inherently different from multi-class classification. 
{
Specifically, since we compute FWT using ROC-AUC as a base metric, the random reference value is $0.5$, whereas, in image classification, it is the ratio between 1 and the number of classes in the single task. This difference exacerbates the complexity of a comparison of FWT in these two settings.
However, we argue that interpreting $FWT$ in this context requires additional in-depth research focused on leveraging concepts similarity for one-class anomaly detection.
}

    \begin{figure}[h!]
        \centering
        \includegraphics[width=0.46\textwidth]{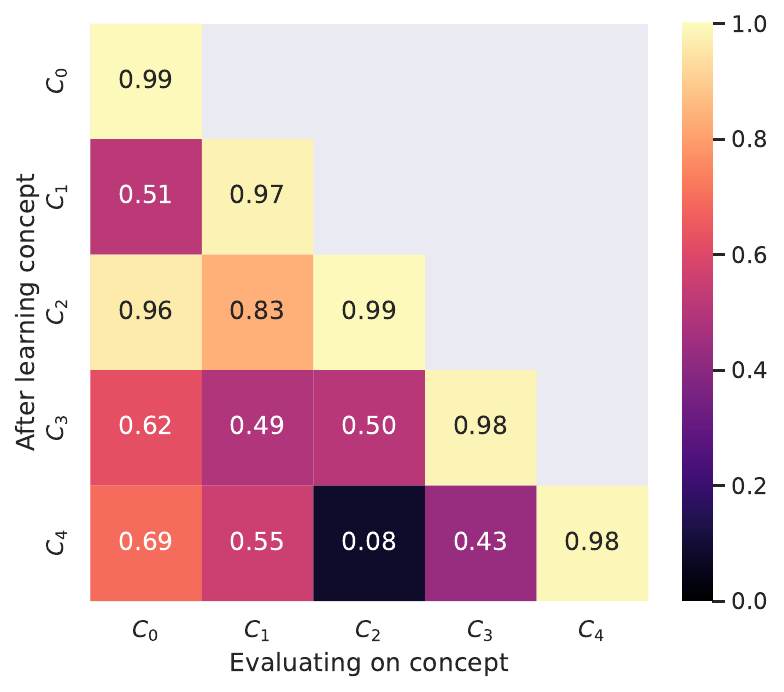}
        \caption{Concept-level ROC-AUC performance for the lifelong learning scenario. Each row $i = 0, 1, \dots $ represents the performance on all concepts observed so far after learning the concept $C_i$ (WIND (R) dataset; Naive lifelong strategy; VAE base model).}
        \label{fig:wind_vae}
    \end{figure} 
    
    \begin{figure}[h!]
        \centering
        \includegraphics[width=0.46\textwidth]{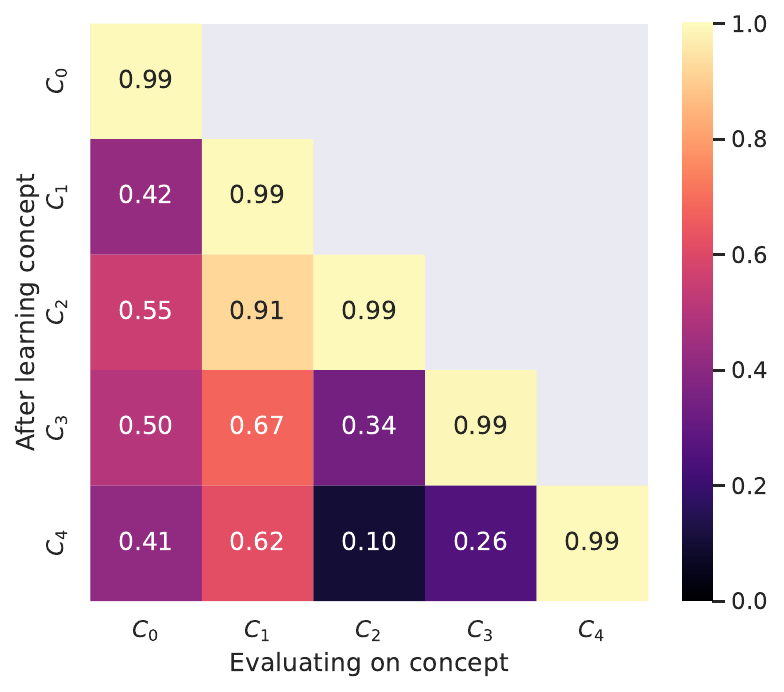}
        \caption{Concept-level ROC-AUC performance for the lifelong learning scenario. Each row $i  = 0, 1, \dots$ represents the performance on all concepts observed so far after learning the concept $C_i$ (WIND (R) dataset; Naive lifelong strategy; LOF base model).}
        \label{fig:wind_LOF}
    \end{figure} 

In summary, {we} observed {a} gap in {ROC-AUC} performance (Naive lifelong $vs.$ MSTE) {alongside with the presence of forgetting (emphasized by negative values of BWT) and degrees of concept similarity (observed via BWT and $FWT$)}. {These phenomena show} that lifelong learning scenarios generated with our approach allow us to adapt non-lifelong datasets to lifelong anomaly detection scenarios, {introducing lifelong challenges {which yield} more demanding conditions for models {(\textbf{RQ1})}. 
{

Additionally, our results show that tackling anomaly detection problems from a lifelong learning perspective can enable a more comprehensive evaluation of models, including measuring the impact of forgetting and task transferability on models. 
%
Concept-level granularity in results also allows us to uncover relationships between concepts in the learning scenario, as well as highlight specific concepts for which models are challenged more than others, enabling a more comprehensive evaluation and in-depth model analysis.
}

\subsection{Impact of the adoption of a replay-based lifelong learning strategy}
{In this subsection, we analyze the impact of the adoption of a lifelong knowledge retention strategy (Replay) on the performance of anomaly detection models. Our effort is aimed at verifying whether the adoption of lifelong learning strategies may be beneficial to improve model performance in a lifelong anomaly detection scenario (\textbf{RQ2})}.

\begin{figure*}[h!]
    \centering
    \includegraphics[width=0.95\textwidth]{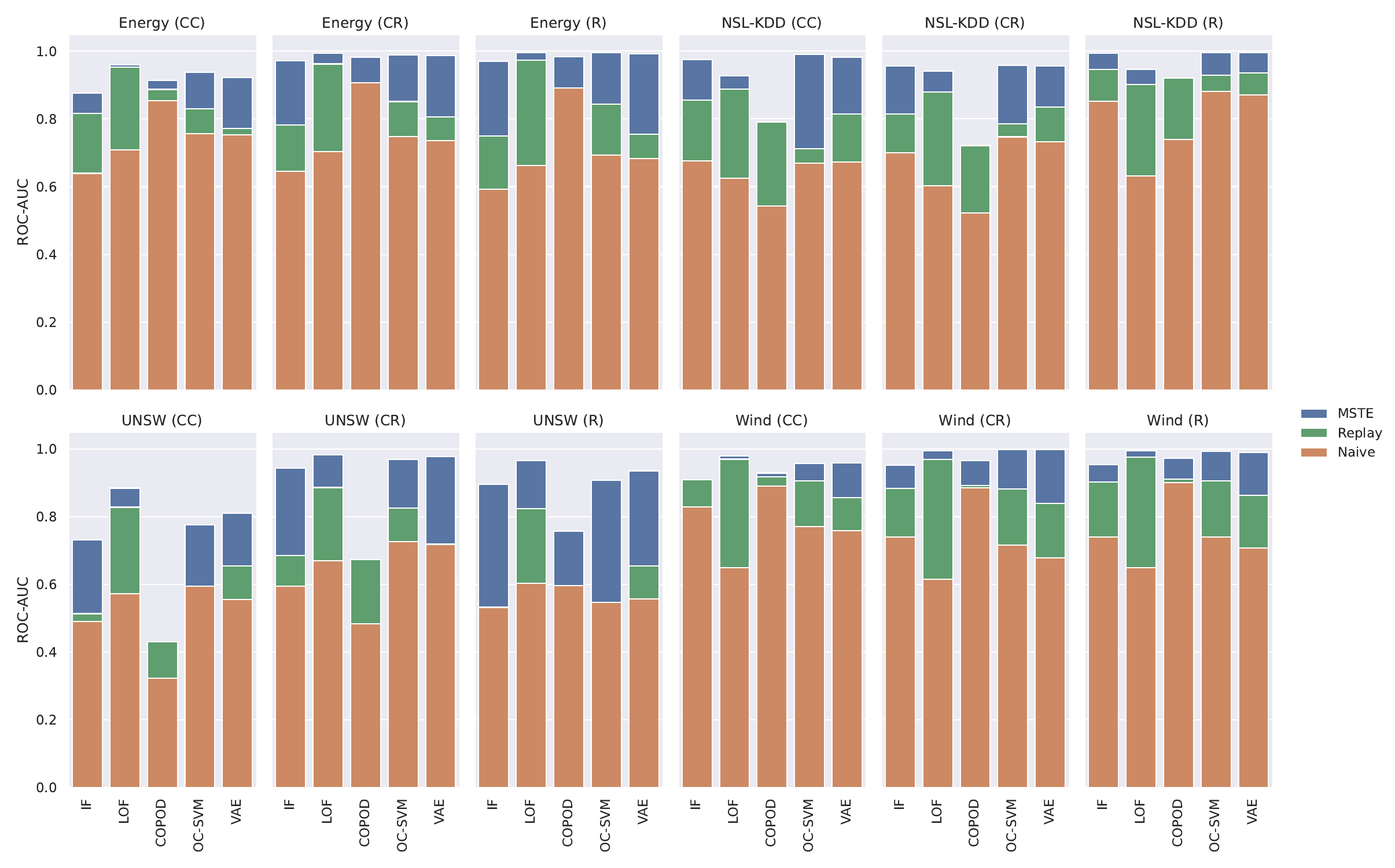}
    \caption{{Summary of experimental results for all datasets (Energy, UNSW, Wind) in three scenario types (CC, CR, R) comparing non-lifelong (Naive), lifelong (Replay), and upper bound (MSTE) learning strategies. This figure illustrates the performance gap between non-lifelong and lifelong strategies in lifelong anomaly detection scenarios.}}
    \label{fig:results-plot}
\end{figure*}



{In order to assess the impact of introducing Replay strategy, we compare anomaly detection performance (ROC-AUC) and backward transfer (BWT) for Replay (Table \mbox{\ref{tab:results-replay}}) and the Naive lifelong strategy (Table \mbox{\ref{tab:results}}).}
We can observe that the Replay strategy brings various levels of improvement{, as shown by higher values of ROC-AUC and BWT}. In some cases, the improvement margin is particularly high. For example, in Energy (CC), the Replay strategy can improve ROC-AUC by $0.18$ (IF).
%
There are also many cases in which the improvement margin is moderate. For example, in Wind (CC), Replay yields a ROC-AUC improvement of $0.08$ (IF).
%
Finally, the improvement margin is quite limited in cases such as UNSW (CC), in which Replay improves the performance of Naive by just $0.02$ (IF).
Similar patterns can be observed across all base models (IF, LOF, COPOD, OC-SVM, VAE).
{A summarized visual perspective of these results is shown in Figure {\ref{fig:results-plot}}, which emphasizes the differences between Naive, Replay, and MSTE.}

Results {in Table {\ref{tab:results-replay}}} show that in the majority of cases (54 out of 60), the simple Replay strategy achieves better results in terms of ROC-AUC than the Naive lifelong approach. 
Most of the exceptions regard the UNSW (R) scenario. 
We attribute this result to the concept complexity in UNSW and the simplicity of the Replay strategy. In this scenario, it is evident that more complex lifelong strategies are required to outperform the Naive baseline. 

The improvement in performance is also supported by the observation of a decrease in forgetting, as shown by improvements in Backward Transfer (BWT) results when comparing Naive lifelong and Replay.
Improvements are considerable, for example, in UNSW (R), where BWT for VAE goes from $-0.46$ to $-0.08$ for VAE, as well as in Wind (CR), where BWT goes from $-0.46$ to $-0.18$.
Improvements with the Replay strategy compared to Naive lifelong can also be observed in Figure \ref{fig:wind_vae_replay} in comparison to Figure \ref{fig:wind_vae}, where we observe a significant increase in performance and a decrease in forgetting in all cases except two (performance on $C_0$ after learning $C_2$, and performance on $C_1$ after learning $C_1$). These two cases highlight that, although the Replay strategy is expected to improve the average model performance due to consideration of all concepts, it also provides additional challenges for the model, which is tasked to learn multiple concepts. As a result, minor decreases in performance for specific concepts should be expected.

We note that there is still a gap between Replay results and simulated upper-bound MSTE results. We can observe that MSTE presents better results than Replay in most cases (56 out of 60). 
%
There are 4 cases in which the Replay strategy is better than MSTE.
The reason behind this phenomenon can be attributed to the fact that while MSTE can build specialized models for each concept, it cannot leverage knowledge from multiple tasks, thus exploiting task similarity, which, in contrast, is supported in a basic form by Replay. 

    \begin{figure}[h!]
        \centering
        \includegraphics[width=0.45\textwidth]{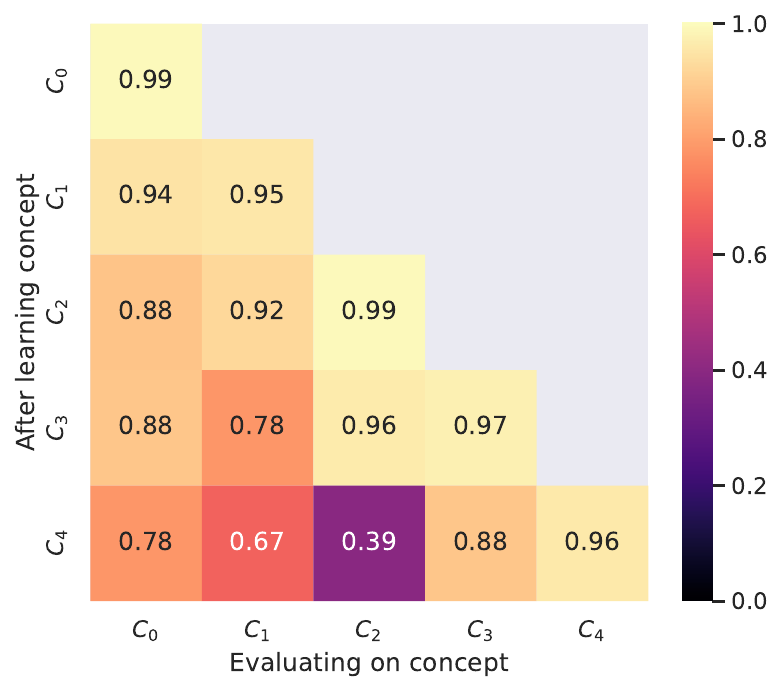}
        \caption{Concept-level ROC-AUC performance for the lifelong learning scenario. Each row $i  = 0, 1, \dots$ represents the performance on all concepts observed so far after learning the concept $C_i$ (WIND (R) dataset; Replay strategy; LOF base model).}
        \label{fig:wind_vae_replay}
    \end{figure} 


{The improvements in ROC-AUC and BWT observed with Replay when compared with Naive}
suggest that adopting lifelong learning techniques and learning strategies referenced in Section \ref{sec:background}, as well as designing new ones, will be beneficial for the advancement of anomaly detection in lifelong learning scenarios {\textbf{(RQ2)}}. 
{At the same time, the discussed gap in performance between Replay and MSTE suggests that more robust lifelong strategies may be devised to further improve model performance in complex lifelong anomaly detection scenarios.}

\subsection{Summary of gained insights}
{In summary, our experimental analysis {discussed above} led us to learn the following insights:}
\begin{itemize}
\item{{A performance gap exists between non-lifelong and lifelong learning strategies in lifelong anomaly detection scenarios {(comparing MSTE vs. Naive)} {(\textbf{RQ1})}. This gap was systematically observed across all scenarios and datasets, and needs to be addressed to increase the reliability of anomaly detection models.}}
\item{{In such scenarios, the adoption of (even simple) lifelong learning strategies such as Replay enables an improvement in anomaly detection performance when compared to non-lifelong models {(comparing Replay vs Naive) {(\textbf{RQ2})}.}}}
\item{{Forgetting was broadly observed for many anomaly detection base models when exposed to the complexity of lifelong scenarios, {as emphasized by negative values of BWT}. This phenomenon suggests that there is an opportunity to build more robust models that simultaneously deal with adaptation and knowledge retention {(\textbf{RQ1, RQ2})}.}}
\item{{Our in-depth lifelong evaluation of model performance revealed cases of positive concept similarity (task similarity). This aspect translates in high values of the forward transfer metric {(FWT)}, leading to above-random anomaly detection performance when models are faced with data from not yet learned concepts. Concept similarity is also leveraged by models to improve performance on previously learned concepts after learning a similar concept, resulting in improved values of backward transfer {(BWT)}.}}
\end{itemize}

\section{Conclusions}
\label{sec:conclusions}
In this paper, we 
{addressed lifelong learning in the context of anomaly detection. In general, our contribution stands in the definition of a common ground for research on this topic and showcasing challenges, perspectives and insights brought by lifelong learning for anomaly detection. 
Specifically, we
}
%
devised a 
learning setting characterization that could be useful to adopt lifelong learning in the context of anomaly detection.
{Moreover, we} designed a procedure for scenario generation that can be used to create lifelong learning scenarios adopting any standard anomaly detection dataset.
Insights from our experiments {revealed that anomaly detection in lifelong learning scenarios is a challenging problem, and that} there is a performance gap between non-lifelong and lifelong learning strategies, indicating that lifelong scenarios are challenging for commonly adopted {non-lifelong} anomaly detection methods. 
{Moreover, we observed} that lifelong learning strategies {such as Replay} have the potential to tackle these challenges.
Overall, we advocate that lifelong learning is essential in anomaly detection to further bring real-life complexity to the experimental setting, providing advantages compared to static and online scenarios currently adopted in the literature.
{We identified a number of domains, such as cybersecurity, human activity, and industrial processes, where such capabilities can be fruitful due to their dynamic characteristics.}
{
We showed that lifelong learning metrics and concept-level performance observed in the learning scenario enable a more detailed model evaluation that uncovers the impact of forgetting and task transferability. 
}

{As we provided an overview of lifelong learning challenges from an anomaly detection perspective, we believe that our work will enable other researchers to take action by working on open problems that are relevant in lifelong anomaly detection. 
{To start with, anomaly detection researchers may adopt the scenarios (e.g., concept-aware, concept-incremental) in their benchmark datasets to expose anomaly detection models to new complexities. Moreover, they can adopt lifelong metrics (Lifelong ROC-AUC, BWT, FWT), as well as the evaluation protocol proposed in this paper. In the future, efforts should be directed at designing or improving lifelong scenarios and metrics, so that they reflect the real-world anomaly detection complexities even better.}
%
%
{Avenues for future research also include the design of different types of lifelong learning strategies beyond replay-based, such as regularization and architectural, which are popular in image classification, and could be tailored to lifelong anomaly detection}.}

\appendix

\section{Hyperparameters of the different base models}
\label{sec:hyperparameters}

{For reproducibility, Table \hbox{\ref{tab:hyperparameters}} shows the five hyperparameter configurations for all models considered in our experiments.}

\begin{table}[h!]
    \centering
    \caption{{Hyperparameters for all the anomaly detection models considered in our experiments. All models are trained and evaluated five times using all hyperparameter values in the sets shown in the table, and the final results are averaged.}}    
    \label{tab:hyperparameters}
    \begin{tabular}{cl}
    \toprule
        VAE & hidden layers size $ \in \{(16, 4, 16); (12, 8, 12); (19, 9, 19);$ \\
         & $(32, 16, 32); (32, 8, 32) \}$ \\
        LOF & n-neighbors $ \in \{ 5; 10; 15; 20; 25 \} $ \\
        IF  & n-estimators $ \in \{ 25; 35; 50; 75; 100 \}$ \\
        OC-SVM & $\nu \in \{ 0.01; 0.02; 0.03; 0.05; 0.1 \}$ \\
          & $\gamma \in \{ 0.01; 0.02; 0.03; 0.05; 0.1 \} $ \\
        COPOD & parameterless \\
    \bottomrule
    \end{tabular}
\end{table}

 \bibliographystyle{unsrt} 
\bibliography{0-refs}

\end{document}